\def\year{2021}\relax
\documentclass[letterpaper]{article} % DO NOT CHANGE THIS
\usepackage{aaai21}  % DO NOT CHANGE THIS
\usepackage{times}  % DO NOT CHANGE THIS
\usepackage{helvet} % DO NOT CHANGE THIS
\usepackage{courier}  % DO NOT CHANGE THIS
\usepackage[hyphens]{url}  % DO NOT CHANGE THIS
\usepackage{graphicx} % DO NOT CHANGE THIS
\usepackage{natbib}  % DO NOT CHANGE THIS AND DO NOT ADD ANY OPTIONS TO IT
\usepackage{caption} % DO NOT CHANGE THIS AND DO NOT ADD ANY OPTIONS TO IT
\usepackage{amsmath}
\title{Cisco at AAAI-CAD21 shared task: Predicting Emphasis in Presentation Slides using Contextualized Embeddings}
\author{
    Sreyan Ghosh\textsuperscript{\rm 1,3\footnote{These authors contributed equally to this work.}},
    Sonal Kumar\textsuperscript{\rm 2\footnotemark[\value{footnote}]},
    Harsh Jalan\textsuperscript{\rm 3\footnotemark[\value{footnote}]},
    Hemant Yadav\textsuperscript{\rm 3},
    Rajiv Ratn Shah\textsuperscript{\rm 3}
    \\
}
\affiliations{
    \textsuperscript{\rm 1}Cisco Systems, Bangalore, India\\
    \textsuperscript{\rm 2}Department of Computer Science and Engineering, CHRIST (Deemed to be University), Bangalore, India\\
    \textsuperscript{\rm 3}MIDAS, IIIT Delhi \\
    \{gsreyan, skbrahee\}@gmail.com, harshjalan27@yahoo.com, \{hemantya, rajivratn\}@iiitd.ac.in\\

}
\begin{document}

\maketitle

\begin{abstract}
This paper describes our proposed system for the AAAI-CAD21 shared task: Predicting Emphasis in Presentation Slides. In this specific task, given the contents of a slide we are asked to predict the degree of emphasis to be laid on each word in the slide. We propose 2 approaches to this problem including a BiLSTM-ELMo approach and a transformers based approach based on RoBERTa and XLNet architectures. We achieve a score of 0.518 on the evaluation leaderboard which ranks us 3\textsuperscript{\rm rd} and 0.543 on the post-evaluation leaderboard which ranks us 1\textsuperscript{\rm st} at the time of writing the paper.
\end{abstract}

\section{Introduction}
Emphasis Selection for written text in visual media from crowdsourced label distributions was first proposed by \citet{shirani2019learning} and then by \citet{shirani2020let} in SemEval-2020 Task 10, Emphasis Selection for Written Text in Visual Media \citep{shirani-etal-2020-semeval}. AAAI-CAD21 shared task: Predicting Emphasis in Presentation Slides \citep{shirani2021CAD21} builds on the same SemEval-2020 Task 10. Presentation slides have become quite common in workplace scenarios and researchers have previously developed resources that guide presenters on the aspects of overall style, color, and font sizes to ensure that the graphical representation of the slide creates an impact on the viewer's mind and the viewer can relate and understand the message that the presenter is trying to relay through the slide. This shared task aims at designing automated approaches to predict which word in the slide should be emphasized (making bold or italics) to improve the visual appeal of the slide. A pictorial example of what this shared task aims to achieve can be seen in Figure 1.

\begin{figure}[h!]
\centering
\includegraphics[width=1\columnwidth]{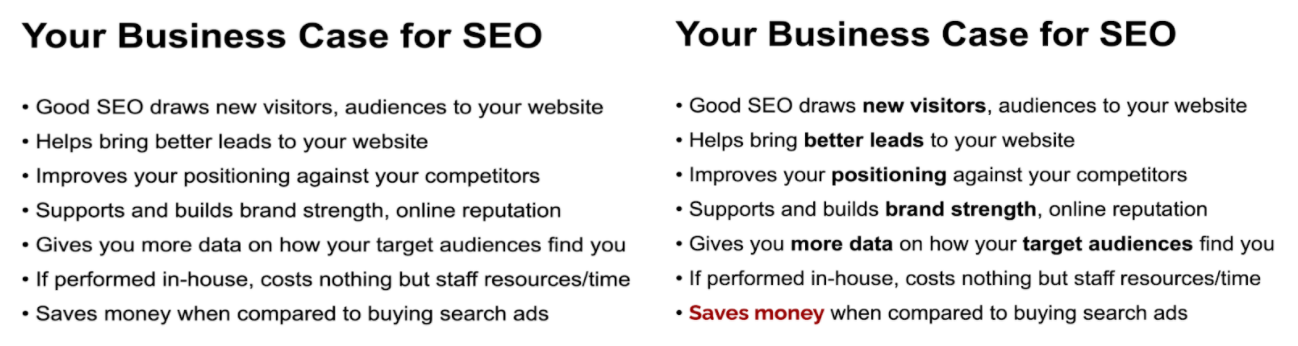} % Reduce the figure size so that it is slightly narrower than the column. Don't use precise values for figure width.This setup will avoid overfull boxes.
\caption{The left slide is plain text. The right side shows the important emphasized words in the slide.}
\label{fig1}
\end{figure}

To solve this problem we treat this task as a sequence labelling problem. Given the contents of an entire slide $d$ = \{$w_1$, $w_2$, ..., $w_n$\} as the input text, we predict the emphasis probability for each word in the contents of the slide $e$ = \{$e_1$, $e_2$, ..., $e_n$\}.

We mainly try two approaches to solve this problem. In our transformers approach, we experiment with two different transformer based model architectures, namely RoBERTa \citep{liu2019RoBERTa} and XLNet \citep{yang2019XLNet}. Our choice of transformer architectures is inspired by the best performing architectures in SemEval-2020 Task 10, Emphasis selection for written text in visual media \citep{singhal2020iitk,anand2020midas}. Both these models were pre-trained on large amounts of unannotated data in an unsupervised manner. A particular token $w$ in a slide is first passed through these transformer models to obtain  embedding of each word in the form of vector representations, after which these vectors are passed through BiLSTM and fully connected layers for classification. We also keep the transformer part of the model trainable and fine-tune the weights on our downstream task of emphasis prediction.

In our second approach, we use a BiLSTM + ELMo  model inspired by the baseline paper \citep{shirani2019learning}. We modify the baseline model and use character embeddings together with pre-trained ELMo embeddings of each word and feed them to BiLSTM + Attention and fully connected layers for predicting emphasis scores. Additionally, we also concatenate some word-level features in the attention output before feeding it to the fully connected layers. A part of our modification is inspired by the team that stood 3rd on the SemEval-2020  Task  10 leaderboard \citep{singhal2020iitk}.

Additionally, we employ 2 approaches to training all models , i.e, BCE or Binary Cross Entropy Loss to directly predict emphasis probabilities and also KL Divergence Loss \citep{kullback1951information} which uses Label Distribution Learning (LDL) \citep{geng2016label} to learn the probabilities of both emphasis and non-emphasis as used in the baseline paper \citep{shirani2019learning}.

\section{Literature Review}
A lot of work in NLP has been done on keyphrase extraction in long texts from scientific articles or news \citep{augenstein2017semeval,zhang2016keyphrase}. Keyword detection mainly focuses on finding important nouns or noun phrases from the input text. Emphasis prediction on the other hand focuses on the automated emphasizing of words in the input text that increase the visual appeal of the text and makes it easier for the viewer of the text to understand the actual message trying to be relayed through it.

Word emphasis prediction has also been explored in spoken data using acoustic and prosodic features \citep{mishra2012word,chen2017automatic}. Emphasis Selection for written text in visual media was first proposed by \citet{shirani2019learning} and then by \citet{shirani2020let} as a SemEval-2020 Task. The baseline paper \citep{shirani2019learning} uses end-to-end label distribution learning (LDL) to predict emphasis scores on short text. The model has an embedding layer which is either Glove  \citep{pennington2014glove}  or ELMo (Peters et al., 2018) followed by BiLSTM + Attention and fully connected layers. They used Adobe Spark Dataset\footnote{https://spark.adobe.com/} for their experiments. Hereon this model will be referred to as the "Baseline" model in our paper.

Team ERNIE \citep{huang2020ernie} from Semeval-2020 Task 10 who stood 1\textsuperscript{\rm st} on the leaderboard, investigated the performance of several transformer-based models including ERNIE 2.0, XLM-RoBERTa, RoBERTa, ALBERT together with a combination of pointwise regression and pairwise ranking loss. The authors also tried some augmentation schemes and word-level lexical features and reported ERNIE 2.0 with the addition of lexical features to be the best performing model on the shared-task dataset.

Team IITK \citep{singhal2020iitk} that stood 3\textsuperscript{\rm rd} on the leaderboard also explored a number of transformer-based datasets including variations of BERT, RoBERTa, XLNet, GPT-2 and XLNet and also a modification on the baseline model. Parts of our modification on the baseline model are also based on this model reported here. Their final results were obtained from a simple ensemble of a number of their transformer-based models.

Team MIDAS \citep{anand2020midas} which stood 11\textsuperscript{\rm th} on the leaderboard also used BERT, RoBERTa and XLNet together with a combination of either BiLSTM and Dense or just Dense layers.

Learning from annotations from different annotators has been explored with majority voting \citep{laws2011active} or by learning individual annotator expertise \citep{10.1145/3178876.3186033,rodrigues2017deep,Rodrigues2013SequenceLW}. Most work on this takes only one label sequence as correct. The baseline paper \citep{shirani2019learning}  was the first work to have used Label Distribution Learning \citep{geng2016label} for a sequence labeling task. We also explore this learning scheme in the experiments mentioned in our paper together with Binary Cross Entropy loss on probabilities obtained from the dataset annotations. 

% An example of the tagging scheme and probability calculation can be seen in table 2.l

\section{Background}
\subsection{Problem Definition}
Given a sequence of words or tokens $d$ = \{$w_1$, $w_2$, ..., $w_n$\} in a slide, the task is to compute a probabilistic score $e_i\:\epsilon\:[0,1]$  for each $w_i$ in $d$ which indicates the degree of emphasis to be laid on the word.

\subsection{Evaluation Metric}
The evaluation metric for our problem is defined as follows:

For a given m (1,5 and 10), we first define 2 sets, $S_m^{(x)}$ - set of $m$ words with top $m$ probabilities according to ground truth and $\hat{S}_m^{(x)}$ - set of $m$ words with top $m$ according to the model predictions. To get $S_m^{(x)}$, each word in the sentence has been manually annotated by 8 annotators. Based on these 2 sets, we define $Match_m$ as:

\begin{equation}
\text {Match}_{m} = \frac{\sum_{x \in D_{\text {test}}}\left|S_{m}^{(x)} \cap \hat{S}_{m}^{(x)}\right| /min(m,|x|)}{\left|D_{\text {test}}\right|}
\end{equation}

where $D_{test}$ is the dataset and $x$ is the token instance. We find $Match_m$ for m $\in \{1,5,10\}$ and express our final score as a simple average over all 3 of them.

\subsection{Dataset}
Dataset Statistics for the dataset provided in the AAAI-CAD21 shared task is shown in Table 1. Each training instance is a complete slide with all the tokens present in the slide. Additionally, the sentence-wise divisions in the slides are also provided in the data. The entire training dataset was annotated by a total of 8 annotators on token-level emphasis. The dataset was annotated with a $BIO$ tagging scheme where each annotator either annotated the token as an emphasized token $(B\:or\:I)$ or not $(O)$. Thus, the probability of emphasis for each token was calculated as an average score of all annotations. The annotation scheme and the emphasis probability calculation has been shown with an example in Table 3. More information about the task and data creation can be found in \citet{shirani2021CAD21}.

\begin{table}[h!]
    \centering
    \begin{tabular}{|l|l|l|l|}
    \hline
         & \textbf{Total Slides} & \textbf{Total Sentences} & \textbf{Total Tokens}  \\
         \hline
         Train & 1241 & 8849 & 96934  \\
         \hline
         Dev & 180 & 1175 & 12822  \\
         \hline
         Test & 355 & 2569 & 28108  \\
         \hline
    \end{tabular}
    \caption{Train, Development and Test Dataset Description}
    \label{table: 1}
\end{table}

\begin{table}[h!]
    \centering
    \begin{tabular}{|l|l|l|l|}
    \hline
       & \textbf{Min} & \textbf{Max} & \textbf{Average} \\
    \hline
    Train & 13 & 180 & 78 \\
    \hline
    Dev & 15 & 164 & 71 \\
    \hline
    Test & 17 & 181 & 79 \\
    \hline
    \end{tabular}
    \caption{Token length description}
    \label{table:2}
\end{table}

\begin{table*}[t]
    \centering
    \begin{tabular}{|c|c c c c c c c c c c|}
    \hline
    \textbf{Words} & \textbf{A1} & \textbf{A2} &\textbf{A3} &\textbf{A4} &\textbf{A5} &\textbf{A6} &\textbf{A7} &\textbf{A8} &\textbf{Freq.[B+I,O]} &\textbf{Probs.} \\
    \hline
    • & O & O & O & O & O & O & O & O & [0,8] & 0.0 \\
    Have & O & O & O & O & O & O & O & O & [0,8] & 0.0 \\ 
    population & O & B & O & O & B & O & O & O & [2,6]  & 0.25 \\
    counts & O & O & O & O & O & O & O & O & [0,8] & 0.0 \\
    for & O & O & O & O & O & O & O & O & [0,8] & 0.0 \\
    three & O & B & O & O & O & O & O & B & [2,6] & 0.25 \\
    key & O & I & O & O & O & O & O & I & [2,6] & 0.25 \\
    species & O & I & B & O & B & O & O & I & [4,4] & 0.5 \\
    \hline
    \end{tabular}
    \caption{Annotation scheme and emphasis probability calculation on a sample sentence from the Train dataset.}
    \label{tab:my_label}
\end{table*}

\section{System Description}
\subsection{Token Level Features}

We tried investigating the data to find token level features that can enhance the performance of our BiLSTM-ELMo model. We tried finding features by analyzing a particular feature's average emphasis score and the number of times a word with that feature occurred in our dataset. The average emphasis scores of the token with these features and the total count can be found in Table 4. Initially, we tried only shape and syntactic features of words by concatenating them with the attention output as described in our system description. The only feature that had given us an improvement over the baseline score was POS (Parts of Speech) tags.

\begin{figure}[h!]
\centering
\includegraphics[width=0.8\columnwidth]{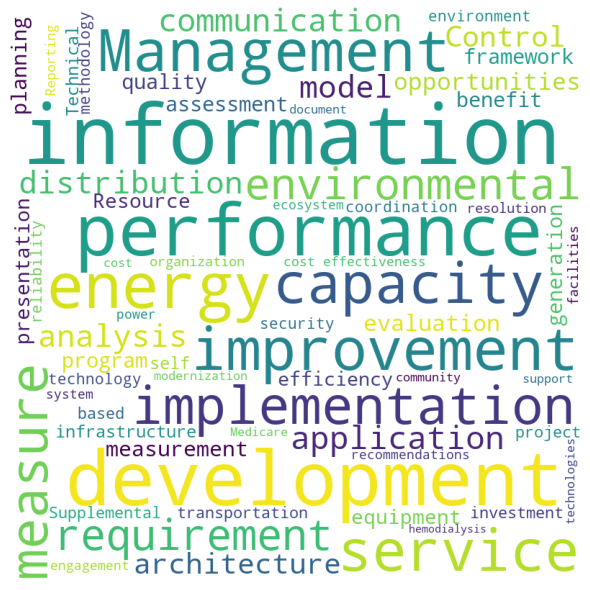} % Reduce the figure size so that it is slightly narrower than the column. Don't use precise values for figure width.This setup will avoid overfull boxes.
\caption{Word Cloud of tokens having emphasis probability of $\geq0.5$}.
\label{fig1}
\end{figure}

\begin{table}[h!]
    \centering
    \begin{tabular}{|l|l|l|}
    \hline
    \textbf{Type} & \textbf{Train (Avg/Nos.)} & \textbf{Dev (Avg/Nos.)} \\
    \hline
    Punctuation & 0.031/14726 & 0.034/2082 \\
    \hline
    UpperCase start & 0.136/21195 & 0.157 \\
    \hline
    Contain numbers & 0.045/2893 & 0.055/308 \\
    \hline
    All Upper Case & 0.092/4498 & 0.116/523 \\
    \hline
    Inside Brackets & 0.002/3598 & 0.009/7560\\
    \hline
    Keyphrase Tags & \textbf{0.25/12723} & \textbf{0.35/1179} \\
    \hline
    Overall & 0.102/96934 & 0119/12822 \\
    \hline
    \end{tabular}
    \caption{Average Emphasis Scores and Count}
    \label{table:3}
\end{table}

\begin{figure*}[t]
\centering
\includegraphics[width=0.9\textwidth]{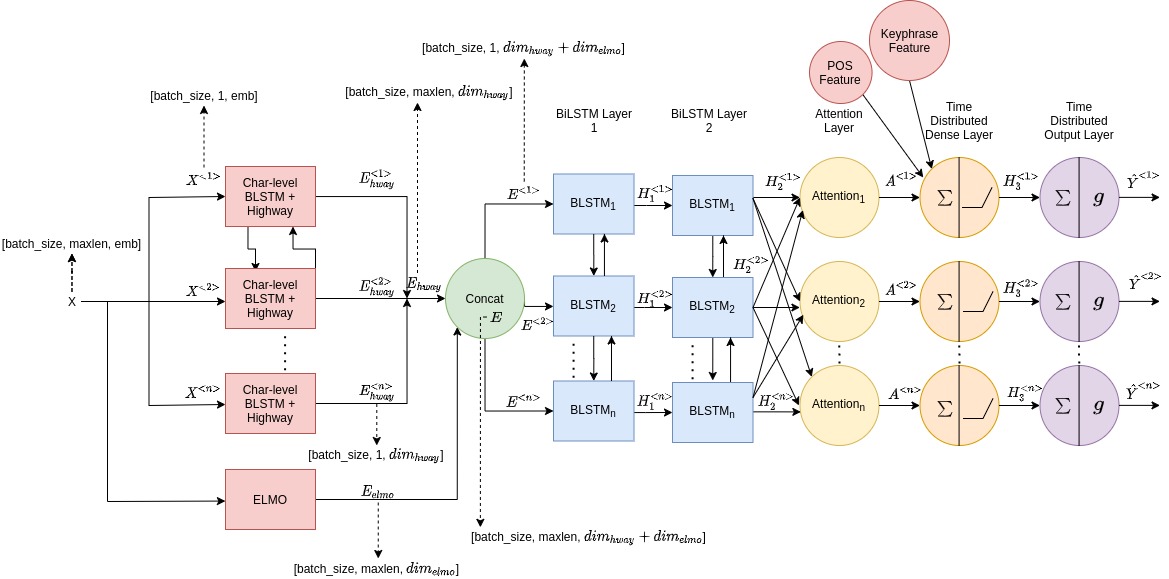} % Reduce the figure size so that it is slightly narrower than the column.
\caption{The BiLSTM-ELMo Model}
\label{fig2}
\end{figure*}
Upon analysis of words with an emphasis score $\geq0.5$ we noticed that most of them were scientific keywords. Thus we created our own feature by training a sequence labeling BiLSTM-CRF model with BERT \citep{DBLP:journals/corr/abs-1802-05365} word embeddings as input to the model with the information extraction datasets used for scientific keyphrase extraction by \citet{sahrawat2019keyphrase} . We use the python flair\footnote{https://github.com/flairNLP/flair} library for this task. The results of the model trained for this task are given in Table 5. The model was trained and inferred with a $BIO$ tagging scheme and it was processed to a binary feature where “B” and “I” tags were termed as 1 and “O” as 0. This feature when used together with POS tags gives us a decent improvement on the baseline results. We name this feature the “Keyphrase Feature” in all our experiments.

\begin{table}[h!]
    \centering
    \begin{tabular}{|l|l|l|l|l|}
    \hline
         & \textbf{Precision} & \textbf{Recall} & \textbf{f1-score} & \textbf{Support} \\
         \hline
     B & 0.6359 & 0.5510 & 0.5904 & 5294 \\
     \hline
     I & 0.6413 & 0.6572 & 0.6492 & 6561 \\
     \hline
     O & 0.9313 & 0.9395 & 0.9354 & 61941 \\
     \hline
     Macro avg & 0.7362 & 0.7159 & 0.7250 & 73796 \\
     \hline
     Weighted avg & 0.8844 & 0.8866 & 0.8852 & 73796 \\
     \hline
    \end{tabular}
    \caption{Keyphrase Extraction Model Results}
    \label{table:4}
\end{table}

\subsection{Our Approach}
\subsubsection{BiLSTM-ELMo Approach}
Our BiLSTM-ELMo approach is inspired by the baseline paper \citep{shirani2019learning} where we extract the ELMo embeddings \citep{peters2018deep} $E_{ELMo}$ for each word in a sequence and additionally, we pass the input through a character-level BiLSTM Network where the combined forward and backward embedding for the last character of each word is then passed through a Highway Layer \citep{singhal2020iitk} which effectively provides us with contextual word-level embeddings $E_{hway}$ for our entire sequence. These contextual word-level embeddings are then concatenated with the extracted ELMo Embeddings for each word to  produce the final word embeddings $E$. 

We pass $E$ through a BiLSTM Layer followed by an Attention Layer. The output of the attention layer is then concatenated with the POS tags \citep{singhal2020iitk} and Keyphrase Feature. for the corresponding word at each time-step. Now combined, the attention output and the external features are fed to a Time Distributed Dense Layer followed by our Time Distributed Output Layer. The activation function $g$ of the Output Layer is either Sigmoid or Softmax depending on whether the Loss Criterion is Binary Cross-Entropy Loss (in case of Sigmoid Activation) or Kullback-Liebler Divergence Loss (in case of Softmax Activation) used for Label Distribution Learning (LDL).

\begin{figure*}[t]
\centering
\includegraphics[width=0.9\textwidth]{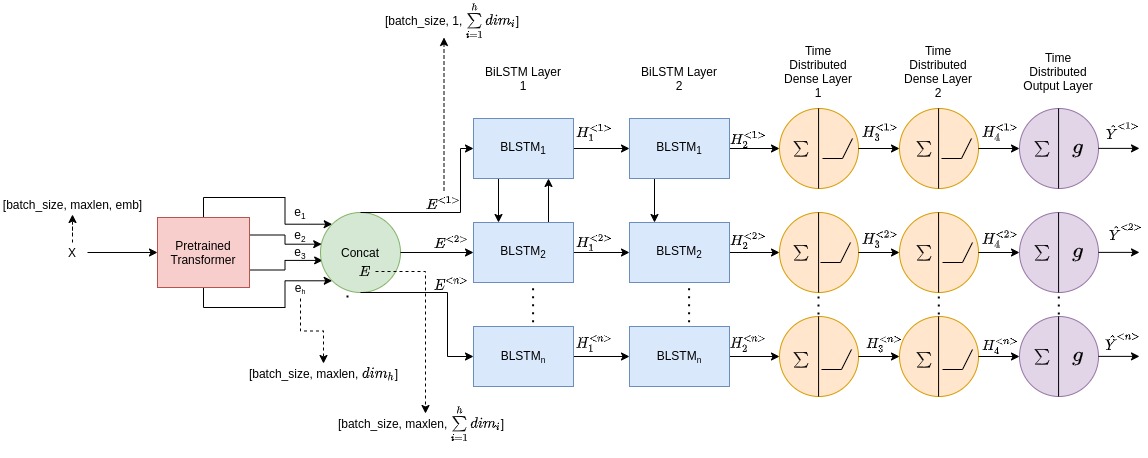} % Reduce the figure size so that it is slightly narrower than the column.
\caption{The Transformer-Based Model}
\label{The Transformer-Based Model}
\end{figure*}

\subsubsection{Transformers Approach}
Our Transformers approach makes use of one of two Transformer Architectures that is, XLNet or RoBERTa. First, the tokenized word input is passed through the transformer architecture and the outputs of all encoding layers of the transformer are concatenated together to get the final embedding $E$ for any given word. This embedding $E$ is now fed through a BiLSTM Layer followed by a set of Time Distributed Dense Layer. Finally, the output of the Time Distributed Dense Layers are fed to our Time Distributed Output Layer with Activation Function $g$. Here, $g$ can either be Sigmoid or Softmax when Loss Function is Binary Cross-Entropy or KL-Divergence respectively.

\section{Experimental Setup}
We use PyTorch \footnote{https://pytorch.org/} Framework for our Deep Learning models along with the Transformer implementations, pre-trained models and, specific tokenizers in the HuggingFace library\footnote{https://huggingface.co/transformers/}.\\

In the BiLSTM-ELMo Approach, we use a hidden size of 300 for the character-level LSTM Layers and on top of that, we use one highway layer which gives us word-level embeddings $E_{hway}$. These embeddings are then concatenated with their corresponding ELMo embeddings $E_{ELMo}$ where the embeddings have 2048 dimensions. This concatenated vector is passed through a BiLSTM Layer with an output size of 512 dimensions in each direction. The Attention Layer uses a self-attention mechanism, the output of the attention mechanism is concatenated with the POS tags and Keyword Feature for each word so that this information can be used by the classifier to make better predictions. The final stage of the classifier consists of a Time Distributed Dense Layer with a hidden size of 20 and ReLU Activation. Finally, the output layer has 1 output neuron if the activation function is Sigmoid and the loss function is Binary Cross-Entropy and 2 output neurons in case of a Softmax activation function and KL-Divergence Loss Function. The dropout layer probabilities were set to 0.3 for all layers to avoid overfitting.

In the transformers approach, we used the RoBERTa and XLNet transformers without freezing any layers of the network and the output of all encoder layers are concatenated to make word-level embeddings $E$. These word embeddings are then passed through two BiLSTM Layers with an output size of 256 dimensions in each direction. The output is then fed to a pair of BiLSTM Layers with 256-dimensional output in both directions. In the classifier, the output of the BiLSTM is fed to a pair of Time Distributed Dense Layers with a hidden layer size of 20 and ReLU Activation and finally to the Time Distributed Output Layer which has either 1 or 2 output neurons depending on whether the activation function used is Sigmoid or Softmax respectively. Dropout Layers with Dropout Probability 0.3 are also added to prevent overfitting. 

When using Sigmoid activation, we aim to predict a single output $\hat{y}^{<t>}$ which represents the probability of emphasis to be laid on the $t^{th}$ token. This probability is used with the Binary Cross-Entropy Loss to train the model. However, in the case of Softmax, we predict a probability distribution $\hat{Y}^{<t>}$ over 2 classes $\{0 = no\:emphasis, 1 = emphasis\}$. This distribution is used with the KL-Divergence Loss function to perform Label Distribution Learning. The equations for $\hat{y}^{<t>}$ and $\hat{Y}^{<t>}$ are as follows:

\begin{equation}
{\hat{y}}^{<t>} = Sigmoid(z^{<t>})
\end{equation}

Where $z^{<t>}$ is the logit of the last output layer for the $t^{th}$ token. 

\begin{equation}
{\hat{Y}}^{<t>} = \{Softmax(z_{0}^{<t>}),Softmax(z_{1}^{<t>})\}
\end{equation}

Where $z^{<t>}_i$ is the logit of the last output layer for the $t^{th}$ token and $i^{th}$ class.\\

In both the Transformers and BiLSTM-ELMo approaches, the Binary Cross-Entropy (BCE) Loss as well as the KL-Divergence (KLD) Loss were used to train the models. The $Match_m$ score is used as an evaluation for all our models. The equations for both the loss functions are as follows:

\begin{align}
BCE(y^{<t>},\hat{y}^{<t>}) = &- y^{<t>}.\log(\hat{y}^{<t>})\notag \\ &- (1-y^{<t>}).\log(1-\hat{y}^{<t>})
\end{align}

Where $y^{<t>}\in\{0,1\}$ is the true label for emphasis laid on each token and $\hat{y}^{<t>}$ is the output of the sigmoid activation for each token.

\begin{equation}
KLD(Y^{<t>}||\hat{Y}^{<t>}) = \sum Y^{<t>}.\log\left(\frac{Y^{<t>}}{\hat{Y}^{<t>}}\right)
\end{equation}

Where $Y^{<t>}$ is the true probability distribution for the emphasis laid on each token and $\hat{Y}^{<t>}$ is the output distribution of the softmax activation for each token.\\

We use the Adam Optimizer for training the models with a learning rate of 1e-4 for the BiLSTM-ELMo model for 100 epochs and 2e-5 for the Transformer-based models for 100 epochs. The training was performed on 1 NVIDIA Titan X GPU. Our code is available on Github\footnote{https://github.com/realsonalkumar/CAD21-AAAI21}.
\section{Results}
In Table 6 we present scores for both our BiLSTM-ELMo and Transformers approach trained on both BCE Loss and KLDivergence Loss for LDL. As we can see in the results, LDL as used by (Shirani et. al 2019) doesn't give a huge improvement over results and at times even diminishes the results.

\begin{table}[h!]
\centering
\begin{tabular}{ |l|l|l| } 
 \hline
 \textbf{Model} & \textbf{Dev} & \textbf{Test} \\
 \hline
 BiLSTM-ELMo (Baseline)  & - & 0.475 \\
 \hline
 BiLSTM-ELMo (POS)  & 0.497 & 0.484 \\ 
 \hline
 BiLSTM-ELMo (POS) (LDL) & 0.501 & 0.506 \\ 
 \hline
 BiLSTM-ELMo (POS, Keyphrase) & 0.515 & 0.496 \\ 
 \hline
 BiLSTM-ELMo (POS, Keyphrase) & 0.504 & 0.501 \\
 (LDL) & & \\
 \hline
 XLNet & \textbf{0.536} & \textbf{0.514} \\ 
 \hline
 XLNet (LDL) & 0.529 & 0.491 \\ 
 \hline
 RoBERTa & 0.51 & 0.485\\
 \hline
 RoBERTa (LDL) & 0.515 & 0.47 \\ 
 \hline
\end{tabular}

\caption{Performance of BiLSTM-ELMo and Transformers approach on development and Test set. The results are expressed in terms of average $Match_m$ for m $\in \{1,5,10\}$. LDL indicates that label distribution learning was employed to train the model with KL-Divergence as the loss function, Binary Cross Entropy otherwise. For BiLSTM-ELMo model the extra features concatenated at the attention layer have been mentioned with each experiment. Baseline. indicates the scores by the baseline model defined by \citet{shirani2019learning}}
\label{table:5}
\end{table}
For our final submissions, we tried an ensemble of scores from different models shown in Table 7. Our best scores on the Evaluation leaderboard were obtained using an ensemble of XLNet and RoBERTa with LDL where we stood 3rd. Meanwhile, our best scores on the Post-Evaluation leaderboard were obtained using an ensemble of XLNet and BiLSTM-ELMo approach with POS tags and Keyphrase Feature where we currently stand 1st on the leaderboard. 

\begin{table}[h!]
    \centering
    \begin{tabular}{|l|l|l|}
    \hline
     \textbf{Model} & \textbf{Dev} & \textbf{Test}  \\
     \hline
     XLNet + RoBERTa (LDL) & 0.547 & 0.518 \\
     \hline
     XLNet + BiLSTM-ELMo (Keyphrase)& 0.538 & 0.532 \\
     \hline
     XLNet + BiLSTM-ELMo (LDL) & \textbf{0.55} & \textbf{0.543} \\
     \hline
    \end{tabular}
    \caption{Performance of different ensemble models}
    \label{table:6}
\end{table}

Additionally, we also ran experiments by dividing the presentations into their constituent sentences in the train and development data. Thus each training instance now corresponds to a particular sentence belonging to a presentation slide in the original corpus. The development set results can be found in Table 8. The evaluation scheme used in this experiment uses the same Match$_m$ as described in the Evaluation Metric section but with  m = \(1,2,3,4\) as used in \citet{shirani2020let}.
\begin{table}[h!]
    \centering
    \begin{tabular}{|l|l|}
    \hline
    \textbf{Model} & \textbf{Dev} \\
    \hline
    XLNet & \textbf{0.758} \\
    \hline
    XLNet (LDL) & 0.757 \\
    \hline
    RoBERTa & 0.743 \\
    \hline
    RoBERTa (LDL) & 0.745 \\
    \hline
    
    BiLSTM-ELMo & 0.751 \\
    \hline
    BiLSTM-ELMo (LDL) & 0.752 \\
    \hline
    
    \end{tabular}
    \caption{Sentence-wise results on the Development set}
    \label{table:7}
\end{table}

\begin{figure}[h!]
    \centering
    \includegraphics[width=0.45\textwidth]{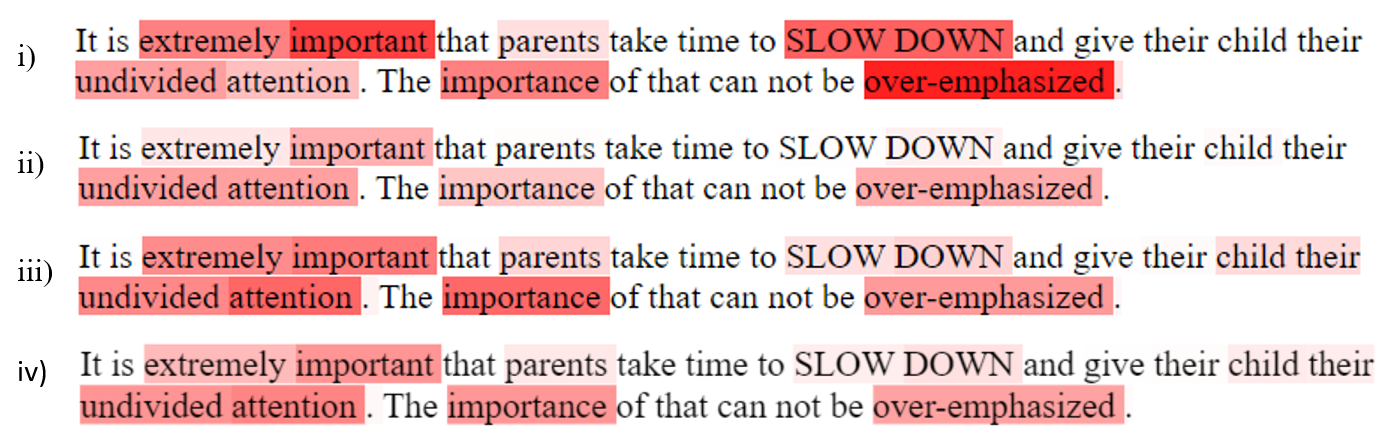}
    \caption{Emphasis Heatmaps i) Ground Truth ii) BiLSTM-ELMo iii) XLNet iv) Best Ensemble Model}
    \label{result_ex}
\end{figure}

\section{Analysis}
\subsection{Length vs Performance}
We wanted to understand how the performance of our models was affected by the length of the instances. Table 9 summarizes the performance of our best performing single model, i.e, XLNet on the development set divided into three sets, Short ($\leq40$ tokens, 80 samples), Medium (40 to 90 tokens, 262 samples), and Long ($>$90 tokens, 50 samples). As we can see, the model performance  deteriorates with the increasing length of the instances. 

\begin{table}[h!]
    \centering
    \begin{tabular}{|l|l|}
    \hline
         & \textbf{XLNet} \\
         \hline
    Small ($\leq40$) & \textbf{0.648}  \\
    \hline
    Medium ($>$40 and $\leq90$) & 0.549 \\
    \hline
    Large ($>$90) & 0.42 \\
    \hline
    \end{tabular}
    \caption{Average $Match_m$ for best performing XLNet model on different size of instances in the development set}
    \label{table:8}
\end{table}

\subsection{Emphasis vs Parts of Speech}
Table 10 shows POS (Parts of Speech) tags vs. average emphasis on the development dataset. We did this experiment to understand how our model predictions performed on each POS tag when compared to the actual human-annotated emphasis scores on the development set. We noticed that the original average emphasis scores were highest on Adjectives followed by Noun. On comparing our models, we found that XLNet was able to almost accurately predict the emphasis scores on Adjectives and Noun respectively, and BiLSTM-ELMo also had the highest predictions on Adjectives and Noun respectively. We also noticed that XLNet did a better job on predicting the emphasis score on different POS tags where the predictions were either very close to the human scores or marginally lesser. On the other hand, we noticed that BiLSTM-ELMo's predictions fell short by bigger margins when compared to XLNet and gave more emphasis to Adverbs than that in the development set. 

\begin{table}[h!]
    \centering
    \begin{tabular}{|l|l|l|l|l|}
    \hline
    \textbf{POS} & \textbf{Count} & \textbf{Human} & \textbf{BiLSTM} & \textbf{XLNet} \\
    \hline
    Noun & 4719 & 0.169 & 0.134 & 0.168 \\
    \hline
    Verb & 1420 & 0.118 & 0.083 & 0.113 \\
    \hline
    Adjectives & 982 & 0.186 & 0.140 & 0.181 \\
    \hline
    Det & 634 & 0.062 & 0.029 & 0.042 \\
    \hline
    Adverbs & 347 & 0.111 & 0.068 & 0.103 \\
    \hline
    Pronouns & 165 & 0.040 & 0.068 & 0.022 \\
    \hline
    Punct & 2082 & 0.034 & 0.015 & 0.025 \\
    \hline
    \end{tabular}
    \caption{POS tags vs. average emphasis on development dataset}
    \label{table:9}
\end{table}

\section{Conclusion}
In this paper, we present our approach to  AAAI-CAD21 shared task: Predicting Emphasis in Presentation Slides. Our best submission gave us an average $Match_m$ of 0.518 placing us 3\textsuperscript{\rm rd} on the Evaluation phase leaderboard and an average $Match_m$ of 0.543 placing us 1\textsuperscript{\rm st} on the Post-Evaluation leaderboard at the time of writing the paper. Future work includes using a hierarchical approach to emphasis prediction as a sequence labeling task using both sentence-level (individual sentence in a slide) and slide-level representations of a word \citep{luo2019hierarchical}.

\section{Acknowledgement}
Rajiv Ratn Shah is partly supported by the Infosys Center for AI at IIIT Delhi. We also thank Sunny Dsouza and Gautam Maurya for their detailed and valuable feedback.

% \section{References}

\bibliography{ex}

\begin{thebibliography}{24}
\providecommand{\natexlab}[1]{#1}
\providecommand{\url}[1]{\texttt{#1}}
\providecommand{\urlprefix}{URL }
\expandafter\ifx\csname urlstyle\endcsname\relax
  \providecommand{\doi}[1]{doi:\discretionary{}{}{}#1}\else
  \providecommand{\doi}{doi:\discretionary{}{}{}\begingroup
  \urlstyle{rm}\Url}\fi

\bibitem[{Anand et~al.(2020)Anand, Gupta, Yadav, Mahata, Gosangi, Zhang, and
  Shah}]{anand2020midas}
Anand, S.; Gupta, P.; Yadav, H.; Mahata, D.; Gosangi, R.; Zhang, H.; and Shah,
  R.~R. 2020.
\newblock MIDAS at SemEval-2020 Task 10: Emphasis Selection using Label
  Distribution Learning and Contextual Embeddings.
\newblock \emph{arXiv preprint arXiv:2009.02619} .

\bibitem[{Augenstein et~al.(2017)Augenstein, Das, Riedel, Vikraman, and
  McCallum}]{augenstein2017semeval}
Augenstein, I.; Das, M.; Riedel, S.; Vikraman, L.; and McCallum, A. 2017.
\newblock Semeval 2017 task 10: Scienceie-extracting keyphrases and relations
  from scientific publications.
\newblock \emph{arXiv preprint arXiv:1704.02853} .

\bibitem[{Chen and Pan(2017)}]{chen2017automatic}
Chen, Y.; and Pan, R. 2017.
\newblock Automatic emphatic information extraction from aligned acoustic data
  and its application on sentence compression.
\newblock In \emph{Thirty-First AAAI Conference on Artificial Intelligence}.

\bibitem[{Geng(2016)}]{geng2016label}
Geng, X. 2016.
\newblock Label distribution learning.
\newblock \emph{IEEE Transactions on Knowledge and Data Engineering} 28(7):
  1734--1748.

\bibitem[{Huang et~al.(2020)Huang, Feng, Su, Chen, Wang, Liu, Ouyang, and
  Sun}]{huang2020ernie}
Huang, Z.; Feng, S.; Su, W.; Chen, X.; Wang, S.; Liu, J.; Ouyang, X.; and Sun,
  Y. 2020.
\newblock ERNIE at SemEval-2020 Task 10: Learning Word Emphasis Selection by
  Pre-trained Language Model.
\newblock \emph{arXiv preprint arXiv:2009.03706} .

\bibitem[{Kullback and Leibler(1951)}]{kullback1951information}
Kullback, S.; and Leibler, R.~A. 1951.
\newblock On information and sufficiency.
\newblock \emph{The annals of mathematical statistics} 22(1): 79--86.

\bibitem[{Laws, Scheible, and Sch{\"u}tze(2011)}]{laws2011active}
Laws, F.; Scheible, C.; and Sch{\"u}tze, H. 2011.
\newblock Active learning with amazon mechanical turk.
\newblock In \emph{Proceedings of the 2011 Conference on Empirical Methods in
  Natural Language Processing}, 1546--1556.

\bibitem[{Liu et~al.(2019)Liu, Ott, Goyal, Du, Joshi, Chen, Levy, Lewis,
  Zettlemoyer, and Stoyanov}]{liu2019RoBERTa}
Liu, Y.; Ott, M.; Goyal, N.; Du, J.; Joshi, M.; Chen, D.; Levy, O.; Lewis, M.;
  Zettlemoyer, L.; and Stoyanov, V. 2019.
\newblock Roberta: A robustly optimized bert pretraining approach.
\newblock \emph{arXiv preprint arXiv:1907.11692} .

\bibitem[{Luo, Xiao, and Zhao(2019)}]{luo2019hierarchical}
Luo, Y.; Xiao, F.; and Zhao, H. 2019.
\newblock Hierarchical Contextualized Representation for Named Entity
  Recognition.

\bibitem[{Mishra, Sridhar, and Conkie(2012)}]{mishra2012word}
Mishra, T.; Sridhar, V.~R.; and Conkie, A. 2012.
\newblock Word prominence detection using robust yet simple prosodic features.
\newblock In \emph{Thirteenth Annual Conference of the International Speech
  Communication Association}.

\bibitem[{Pennington, Socher, and Manning(2014)}]{pennington2014glove}
Pennington, J.; Socher, R.; and Manning, C.~D. 2014.
\newblock Glove: Global vectors for word representation.
\newblock In \emph{Proceedings of the 2014 conference on empirical methods in
  natural language processing (EMNLP)}, 1532--1543.

\bibitem[{Peters et~al.(2018{\natexlab{a}})Peters, Neumann, Iyyer, Gardner,
  Clark, Lee, and Zettlemoyer}]{DBLP:journals/corr/abs-1802-05365}
Peters, M.~E.; Neumann, M.; Iyyer, M.; Gardner, M.; Clark, C.; Lee, K.; and
  Zettlemoyer, L. 2018{\natexlab{a}}.
\newblock Deep contextualized word representations.
\newblock \emph{CoRR} abs/1802.05365.
\newblock \urlprefix\url{http://arxiv.org/abs/1802.05365}.

\bibitem[{Peters et~al.(2018{\natexlab{b}})Peters, Neumann, Iyyer, Gardner,
  Clark, Lee, and Zettlemoyer}]{peters2018deep}
Peters, M.~E.; Neumann, M.; Iyyer, M.; Gardner, M.; Clark, C.; Lee, K.; and
  Zettlemoyer, L. 2018{\natexlab{b}}.
\newblock Deep contextualized word representations.
\newblock \emph{arXiv preprint arXiv:1802.05365} .

\bibitem[{Rodrigues and Pereira(2017)}]{rodrigues2017deep}
Rodrigues, F.; and Pereira, F. 2017.
\newblock Deep learning from crowds.

\bibitem[{Rodrigues, Pereira, and Ribeiro(2013)}]{Rodrigues2013SequenceLW}
Rodrigues, F.; Pereira, F.~C.; and Ribeiro, B. 2013.
\newblock Sequence labeling with multiple annotators.
\newblock \emph{Machine Learning} 95: 165--181.

\bibitem[{Sahrawat et~al.(2019)Sahrawat, Mahata, Kulkarni, Zhang, Gosangi,
  Stent, Sharma, Kumar, Shah, and Zimmermann}]{sahrawat2019keyphrase}
Sahrawat, D.; Mahata, D.; Kulkarni, M.; Zhang, H.; Gosangi, R.; Stent, A.;
  Sharma, A.; Kumar, Y.; Shah, R.~R.; and Zimmermann, R. 2019.
\newblock Keyphrase Extraction from Scholarly Articles as Sequence Labeling
  using Contextualized Embeddings.
\newblock \emph{arXiv preprint arXiv:1910.08840} .

\bibitem[{Shirani et~al.(2019)Shirani, Dernoncourt, Asente, Lipka, Kim,
  Echevarria, and Solorio}]{shirani2019learning}
Shirani, A.; Dernoncourt, F.; Asente, P.; Lipka, N.; Kim, S.; Echevarria, J.;
  and Solorio, T. 2019.
\newblock Learning emphasis selection for written text in visual media from
  crowd-sourced label distributions.
\newblock In \emph{Proceedings of the 57th Annual Meeting of the Association
  for Computational Linguistics}, 1167--1172.

\bibitem[{Shirani et~al.(2020{\natexlab{a}})Shirani, Dernoncourt, Echevarria,
  Asente, Lipka, and Solorio}]{shirani2020let}
Shirani, A.; Dernoncourt, F.; Echevarria, J.; Asente, P.; Lipka, N.; and
  Solorio, T. 2020{\natexlab{a}}.
\newblock Let Me Choose: From Verbal Context to Font Selection.
\newblock \emph{arXiv preprint arXiv:2005.01151} .

\bibitem[{Shirani et~al.(2020{\natexlab{b}})Shirani, Dernoncourt, Lipka,
  Asente, Echevarria, and Solorio}]{shirani-etal-2020-semeval}
Shirani, A.; Dernoncourt, F.; Lipka, N.; Asente, P.; Echevarria, J.; and
  Solorio, T. 2020{\natexlab{b}}.
\newblock {S}em{E}val-2020 Task 10: Emphasis Selection for Written Text in
  Visual Media.
\newblock In \emph{Proceedings of the Fourteenth Workshop on Semantic
  Evaluation}, 1360--1370. Barcelona (online): International Committee for
  Computational Linguistics.
\newblock \urlprefix\url{https://www.aclweb.org/anthology/2020.semeval-1.184}.

\bibitem[{Shirani et~al.(2021)Shirani, Tran, Trinh, Dernoncourt, Lipka, Asente,
  Echevarria, and Solorio}]{shirani2021CAD21}
Shirani, A.; Tran, G.; Trinh, H.; Dernoncourt, F.; Lipka, N.; Asente, P.;
  Echevarria, J.; and Solorio, T. 2021.
\newblock Learning to Emphasize: Dataset and Shared Task Models for Selecting
  Emphasis in Presentation Slides.
\newblock In \emph{Proceedings of CAD21 workshop at the Thirty-fifth AAAI
  Conference on Artificial Intelligence (AAAI-21)}.

\bibitem[{Singhal et~al.(2020)Singhal, Dhull, Agarwal, and
  Modi}]{singhal2020iitk}
Singhal, V.; Dhull, S.; Agarwal, R.; and Modi, A. 2020.
\newblock IITK at SemEval-2020 Task 10: Transformers for Emphasis Selection.
\newblock \emph{arXiv preprint arXiv:2007.10820} .

\bibitem[{Yang et~al.(2018)Yang, Drake, Damianou, and
  Maarek}]{10.1145/3178876.3186033}
Yang, J.; Drake, T.; Damianou, A.; and Maarek, Y. 2018.
\newblock Leveraging Crowdsourcing Data for Deep Active Learning An
  Application: Learning Intents in Alexa.
\newblock In \emph{Proceedings of the 2018 World Wide Web Conference}, WWW '18,
  23–32. Republic and Canton of Geneva, CHE: International World Wide Web
  Conferences Steering Committee.
\newblock ISBN 9781450356398.
\newblock \doi{10.1145/3178876.3186033}.
\newblock \urlprefix\url{https://doi.org/10.1145/3178876.3186033}.

\bibitem[{Yang et~al.(2019)Yang, Dai, Yang, Carbonell, Salakhutdinov, and
  Le}]{yang2019XLNet}
Yang, Z.; Dai, Z.; Yang, Y.; Carbonell, J.; Salakhutdinov, R.~R.; and Le, Q.~V.
  2019.
\newblock Xlnet: Generalized autoregressive pretraining for language
  understanding.
\newblock In \emph{Advances in neural information processing systems},
  5753--5763.

\bibitem[{Zhang et~al.(2016)Zhang, Wang, Gong, and Huang}]{zhang2016keyphrase}
Zhang, Q.; Wang, Y.; Gong, Y.; and Huang, X.-J. 2016.
\newblock Keyphrase extraction using deep recurrent neural networks on twitter.
\newblock In \emph{Proceedings of the 2016 conference on empirical methods in
  natural language processing}, 836--845.

\end{thebibliography}

\end{document}